\documentclass[10pt,journal,final]{IEEEtran}

\usepackage{cite}
\usepackage{graphicx}
\usepackage{bm}
\usepackage{epsfig}
\usepackage{amsmath}
\usepackage{array}
\usepackage{multirow}
\usepackage{color}
\usepackage{amsmath}
\usepackage{xcolor}

\usepackage{amsfonts}
\usepackage{booktabs}
\pdfoutput=1
\usepackage{balance}

\ifCLASSINFOpdf
\else
\fi

\begin{document}
%
\title{{Quality-aware Part Models for Occluded Person Re-identification}}

\author{Pengfei Wang,~\IEEEmembership{Student Member,~IEEE,}
        Changxing Ding,~\IEEEmembership{Member,~IEEE,}
       Zhiyin Shao, \\
       Zhibin Hong,
       Shengli Zhang,~\IEEEmembership{Member,~IEEE,}
       and Dacheng Tao,~\IEEEmembership{Fellow,~IEEE}

\thanks{Pengfei Wang, Changxing Ding, and Zhiyin Shao are with the School of Electronic and Information Engineering,
South China University of Technology,
381 Wushan Road, Tianhe District, Guangzhou 510000, P.R. China
(e-mail: eepengfei.wang@mail.scut.edu.cn; chxding@scut.edu.cn; eezyshao@mail.scut.edu.cn).}
\thanks{Zhibin Hong is with department of Computer Vision
Technology (VIS), Baidu Inc, ShenZhen 518000, China
(e-mail: hongzhibin@baidu.com).}
\thanks{Shengli Zhang is with the College of Electronic and Information Engineering, Shenzhen University, Shenzhen 518052, China
(e-mail: zsl@szu.edu.cn).}
\thanks{Dacheng Tao is with JD Explore Academy, JD.com, Beijing 100176, China
(e-mail: taodacheng@jd.com).}}

\maketitle

\begin{abstract}
 Occlusion poses a major challenge for person re-identification (ReID). Existing approaches typically rely on outside tools to infer visible body parts, which may be suboptimal in terms of both computational efficiency and ReID accuracy. In particular, they  may  fail  when  facing  complex  occlusions,  such as those between pedestrians.
 Accordingly, in this paper, we propose a
novel method named Quality-aware Part Models (QPM) for occlusion-robust ReID.
 First, we propose to jointly learn part features and predict part quality scores.
 As no quality annotation is available, we introduce a strategy that automatically assigns low scores to occluded body parts, thereby weakening the impact of occluded body parts on ReID results.
Second, based on the predicted part quality scores, we propose
a novel identity-aware spatial attention (ISA) module.
In this module, a coarse identity-aware feature is utilized to highlight pixels of the target  pedestrian, so as to handle the occlusion between pedestrians.
Third, we design an adaptive and efficient approach for generating global features from common non-occluded regions with respect to each image pair.
This design is crucial, but is often ignored by existing methods.
QPM has three key advantages: 1) it does not rely on any outside tools in either the training or inference stages; 2) it handles occlusions caused by both objects and other pedestrians;
3) it is highly computationally efficient.
Experimental results on four popular databases for occluded ReID demonstrate that QPM consistently outperforms state-of-the-art methods by significant margins. The code of QPM will be released.
\end{abstract}


\IEEEpeerreviewmaketitle

\section{Introduction}
 \label{Introduction}
\IEEEPARstart{P}{erson} re-identification (ReID) involves spotting a specific person of interest, e.g. a missing child, across disjoint camera views.
Due to the widespread deployment of modern surveillance systems, ReID has attracted increasing attention from both academia and industry \cite{11_zhang2019densely, 06_li2018harmonious, sun2018beyond, 18_xu2018attention, 33chen2017person, 31song2018mask, 21_tian2018eliminating, 22_Robust2017tip,luo2019strong,scheirer2016report, wei2018glad,yan2021beyond,jiang2021ph}.
Most existing ReID approaches assume that the pedestrian's entire body is visible, and tend to ignore the more challenging occlusion situations.
However, in real-world applications, pedestrians are very often occluded by objects or other pedestrians.

\begin{figure}
  {\centerline{\includegraphics[width=0.5\textwidth]{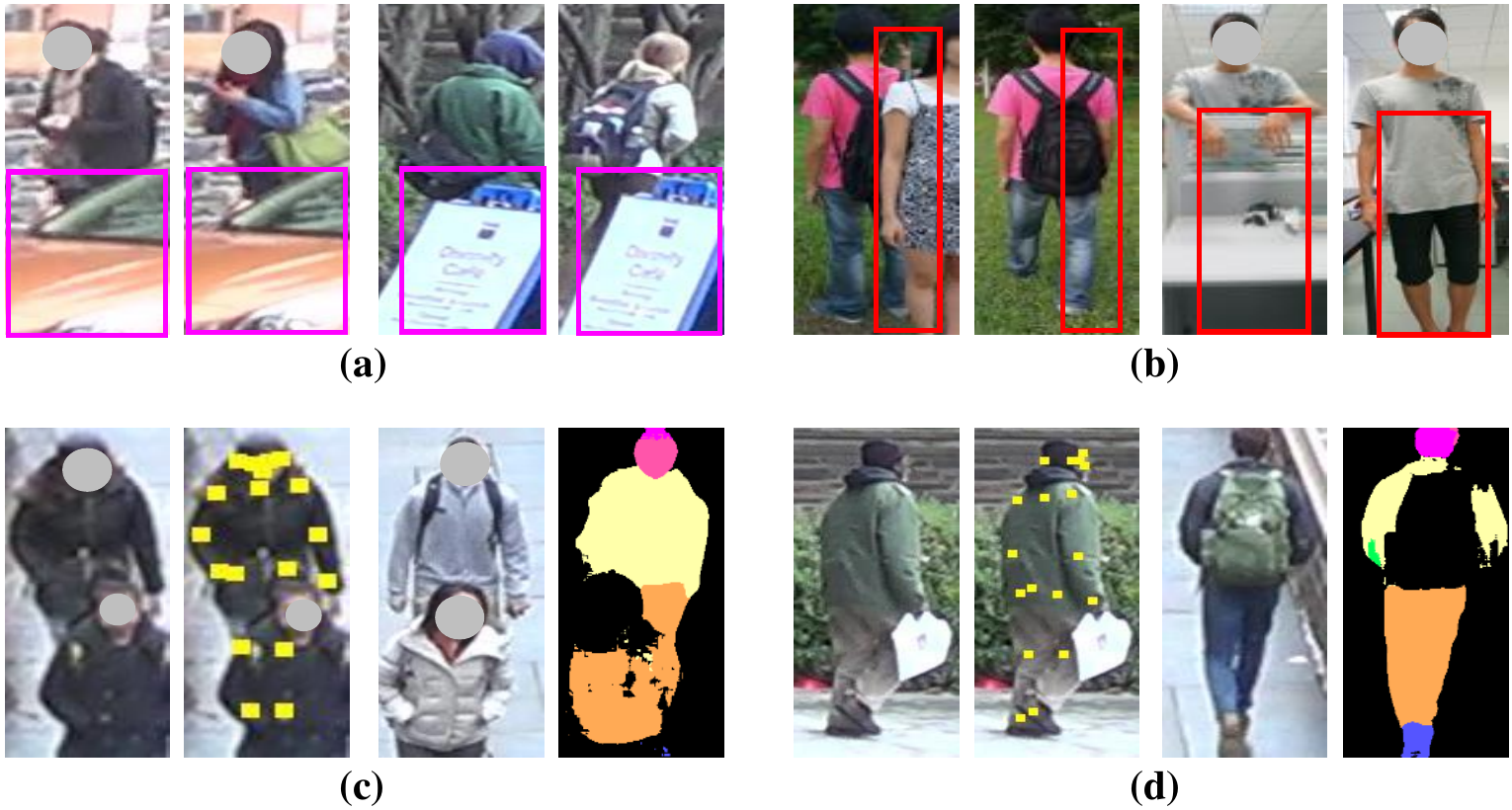}}
  \caption{Example images that illustrate the challenges for occluded ReID: (a) Similar occlusion reduces inter-class distance.
     (b) Dissimilar occlusion enlarges intra-class variation.
     (c) Both human parsing and pose estimation tools  may fail when facing complex occlusions, e.g. occlusion between pedestrians. (d) Both tools
may ignore discriminative accessories, such as backpacks or handbags. (Best viewed in color.)}
  \label{examples1}
  }
\end{figure}

Occlusion poses a major challenge for ReID, as it affects the appearance of pedestrian.
As illustrated in Fig.~\ref{examples1}(a), similar occlusion reduces inter-class distance, which indicates that images of different identities may have similar visual features.
Moreover, as shown in Fig.~\ref{examples1}(b), different occlusions enlarge intra-class distance, meaning that two images of the same pedestrian may be quite different in terms of their appearance.
This is because occlusions may differ as regards their location and content; therefore, occlusion tends to result in incorrect retrieval results.

Since occlusion changes pedestrian appearance, one intuitive solution is to only utilize visible body parts for ReID. Most existing methods adopt this strategy~\cite{miao2019PGFA, Wang_2020_CVPR, Gao_2020_CVPR}.
However, they usually rely on outside tools to acquire the visibility cues of body parts (e.g. the prediction confidence of pose estimation models).
Beside extra computational cost, this strategy may not be robust to complex occlusions, such as occlusions between pedestrians. As illustrated in Fig.~\ref{examples1}(c), both human parsing and pose estimation tools may fail when facing occlusions between pedestrians.
Moreover, visibility is not equivalent to discriminative power.
On the one hand, one visible part may look quite similar across different pedestrians.
For example, nearly all pedestrians' forearms are not covered by any clothes in images captured in summer.
On the other hand, invisible body parts may be occluded by discriminative accessories, such as backpacks and bags, as shown in Fig.~\ref{examples1}(d). These accessories are critical for ReID, but   tend to be ignored by outside tools.
It is therefore reasonable to seek out a robust and easy-to-use method that automatically infers and utilizes discriminative body parts to handle the occlusion problem.

Accordingly, in this paper, we propose a novel framework named QPM for occluded ReID.
QPM includes a part branch and a global branch.
The part branch automatically infers part-specific quality scores rather than visibility scores.
More specifically, it jointly learns discriminative part features and predicts part quality scores in an end-to-end fashion.
This is achieved by including both pair-wise part distances and pair-wise part quality scores in the triplet loss \cite{39hermans2017defense}; as a result, the part branch can automatically assign low quality scores to poor-quality body parts in order to weaken their influence.
Another key benefit of our approach is that it is independent from any outside tools and does not require annotations of quality scores. 
However, when a pedestrian is occluded by other persons, the model may predict a relatively high score for occluded regions; this is because the model cannot well differentiate body parts of different pedestrians. We solve this problem in the global branch.

The global branch includes two main components. First, we propose an identity-aware spatial attention (ISA) module based on the predicted part quality scores.
In this module, a coarse identity-aware feature is processed by a simple two-layer network and optimized using the cross-entropy loss function.
Then, it is utilized to suppress noisy responses and highlight responses from the body region of the pedestrian needing to be identified. Therefore, it can be used to handle occlusions between pedestrians.
Second, we design an adaptive and efficient approach to generate global features from the common non-occluded regions with respect to each image pair.
By contrast, existing works typically extract fixed global features, ignoring the difference in occlusion locations between a pair of images.

In the inference stage, both the part and global features are utilized to calculate the similarity score between each pair of images.
The weighted average of both scores represents the overall similarity of an image pair. We conduct extensive experiments on four popular datasets for occluded ReID, i.e. Partial-iLIDS~\cite{he2018deep}, Partial-REID~\cite{zheng2015partial}, Occluded-Duke~\cite{miao2019PGFA}, and P-DukeMTMC~\cite{zhuo2018occluded}. The results show that our simple QPM model consistently outperforms existing approaches by significant margins.
Moreover, our approach enjoys further advantages of being robust and easy to use.

In conclusion, the main contributions of this paper are summarized as following:
       \begin{itemize}
	
		\item 
	We propose an end-to-end framework that jointly learns discriminative features and predicts part quality scores. Compared with existing works, it does not rely on any outside tools in either the training or inference stages.

		\item 
		We propose a novel identity-aware spatial attention (ISA) approach that efficiently handles the occlusion between pedestrians. Experimental results prove that it outperforms existing spatial attention methods.

		\item 
		We introduce an Adaptive Global Feature Extraction (AGFE) module that extracts global features from the commonly non-occluded regions for each image pair, which significantly promotes ReID performance.

	\end{itemize}

The remainder of this paper is organized as follows. We first review the related works in Section II. Then, we describe the proposed QPM in more detail in Section III. Extensive experimental results on three benchmarks are reported and analyzed in Section IV, after which the conclusions of the present work are outlined in Section V.

\section{Related Work}
\subsection{Occluded Person ReID Models}
One main challenge for occluded ReID is to identify visible body regions.
Most existing works utilize visibility cues provided by outside tools~\cite{59kalayeh2018human,qi2018maskreid,31song2018mask,10_zheng2019Pose,liuCVPR2018pose,08_su2017pose}. Here, we divide occluded ReID methods into two categories depending on whether or not outside tools are required during training and testing.

The first category of methods employs outside tools in both the training and testing stages~\cite{miao2019PGFA,Wang_2020_CVPR,Gao_2020_CVPR}.
For example,
Miao \textit{et al.}~\cite{miao2019PGFA} utilized pose landmarks to identify visible local patches and only adopt commonly visible local patches of one image pair for matching.
 Wang \textit{et al.}~\cite{Wang_2020_CVPR} utilized pose landmarks to learn high-order relation and topology information of the visible local features, so as to better match probe with gallery images.
 Gao \textit{et al.}~\cite{Gao_2020_CVPR} employed graph matching and utilized pose landmarks to self-mine part visibility scores. They then match probe and gallery images by calculating the part-to-part distances in visible regions.
However, in addition to its extra computational cost, another key downside of this approach is that external tools may not be reliable when encountering complex occlusions, as illustrated in Fig.~\ref{examples1}(c-d).

The second category of methods avoids using outside tools in the testing stage \cite{he2019foreground-aware,he2020guided,he2018recognizing,zhuo2019novel}.
For example, under the guidance of human masks, He \textit{et al.}~\cite{he2019foreground-aware} designed an occlusion-sensitive foreground probability generator that enables the model to focus on non-occluded human body parts.
He \textit{et al.}~\cite{he2020guided} further combined pose landmarks and human masks to generate spatial attention maps that guide discriminative feature learning.
These approaches can reduce the impact of occlusions on the extracted features.
Despite the convenience this affords during testing, this approach still relies on the visibility information of body part for each training image.

\begin{figure*}[t]
  \centerline{\includegraphics[width=1.0\textwidth]{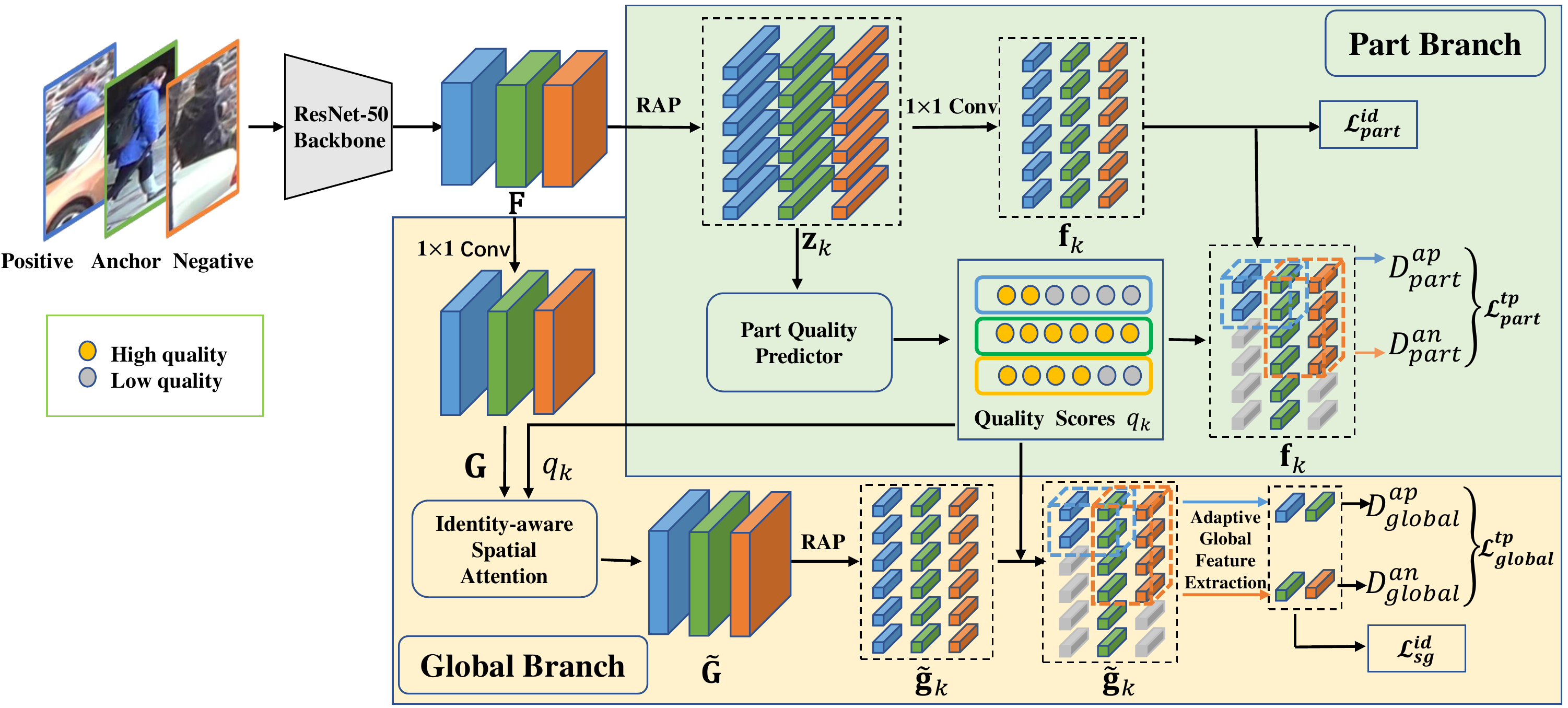}}
  \caption{
Model architecture of QPM in the training stage.
Based on ResNet-50,
QPM builds a part branch and a global branch.
The part branch jointly extracts part features and predicts part quality scores. The quality scores are utilized to weaken the impact of occluded body parts. The global branch first adopts a novel Identity-aware Spatial Attention module to handle occlusions between pedestrians. Then, it adaptively and efficiently extracts global features from the common non-occluded regions for each image pair. In the testing stage, the global features, part features, and part quality scores are utilized
together for occluded ReID. (Best viewed in color.)
}
  \label{net}
\end{figure*}

Moreover, two very recent works \cite{zhu2020identity,sun2019perceive} try to predict visible scores without outside tools. VPM \cite{sun2019perceive} 
classifies each pixel in the image into one body part.
However, it is designed for the partial ReID problem and cannot be directly used to solve the occluded ReID problem. Specifically, VPM classifies each pixel in the partial image into one body part, assuming there is no occlusion in the partial image. ISP \cite{zhu2020identity} performs cascaded clustering on feature maps to generate the pseudo-labels of human parts for each pixel. However, ISP contains no strategy to handle the occlusion problem between pedestrians. In addition, the clustering process is time-consuming.

In a departure from existing works, our approach aims to predict the part quality rather than visibility in an end-to-end framework. And we do not employ any outside tools or require any quality annotations in either the training or inference stages.  Moreover, our approach is robust to complex occlusions. In conclusion, QPM is a powerful and efficient model.

\subsection{Part-based Person ReID Models}
Due to their powerful representation ability, part-based methods are popular for ReID. Depending on the way to obtain body part locations, we divide existing works into three categories.

\noindent\textbf{Fixed Location-based Methods.}
Methods in this category typically split the output feature maps of one backbone model into several stripes in fixed locations~\cite{sun2018beyond,03_wang2018learning,02_zheng2019pyramidal}. Part features are then respectively extracted from the stripes.  For
example, Sun \textit{et al.}~\cite{sun2018beyond} uniformly divide the output feature maps into 6 horizontal stripes to represent different part-level features. Wang \textit{et al.} ~\cite{03_wang2018learning} also partition one image into
horizontal stripes. The main advantage of this strategy lies in its efficiency.

\noindent\textbf{Outside Tools-based Methods.}
These methods utilize outside tools, e.g. pose estimation \cite{09_Saquib2018pose, 08_su2017pose, 10_zheng2019Pose, 04_zhao2017spindle} and human parsing models \cite{59kalayeh2018human, 31song2018mask}, to detect body parts. They then extract part features from the detected body parts. There are two key downsides of these approaches: first, they require additional computational cost; second, ReID  performance is vulnerable to the reliability of outside tools.

\noindent\textbf{Attention-based Methods.}
These methods predict body part locations based on the feature maps produced by ReID models \cite{43zhao2017deeply,06_li2018harmonious,07_li2017learning,gong2021lag}. For example, Zhao \textit{et al.}~\cite{43zhao2017deeply} proposed to predict a set of masks and perform element-wise multiplication between one mask and each channel of the feature maps to produce part-specific features. In comparison, Li \textit{et al.}  \cite{06_li2018harmonious} designed a hard regional attention model that can predict
bounding boxes for each body part.  However, the lack
of explicit supervision for part alignment may cause difficulty
in the optimization of attention models.

In part-based ReID methods, it is a common practice to concatenate part features as the final representation \cite{10_zheng2019Pose,sun2018beyond,02_zheng2019pyramidal,43zhao2017deeply,li2021diverse,wan2019concentrated,ding2020multi,wang2021batch}. However, it is less effective for occluded ReID, as it ignores the impact of features from occluded parts. In this paper, we accordingly propose to jointly learn part features and predict part quality. For simplicity, we extract part features from fixed part locations. If part locations are provided by outside tools or attention modules, the performance of our approach can be further promoted.

\section{Methods}

The architecture of QPM is illustrated in Fig.~\ref{net}. It consists of a part feature learning branch and a global feature learning branch. The part branch outputs $K$ part-level features, as well as $K$ quality scores that indicate the discriminative power of each body part.
The global branch generates global features that are adaptively and efficiently extracted from the common non-occluded regions for each gallery-query image pair.
In the following, we will introduce three key designs in the two branches individually.

\subsection{Joint Learning Part Feature and Quality Scores}
\noindent\textbf{Part Feature Extractor.}
Following \cite{sun2018beyond}, we adopt ResNet-50 as backbone and remove its last spatial down-sampling operation to increase the size of the output feature maps. The output feature maps are denoted as $\mathbf{F}$ for simplicity. To obtain the part features, $\mathbf{F}$ is first uniformly split into $K$ parts in the vertical orientation.
Following \cite{sun2018beyond}, we set $K$ as 6 in this work.
Next, the feature maps for each part are processed by a Region Average Pooling (RAP) operation and one $1\times 1$ Conv layer. The parameters of the  Conv layer are not shared between parts. For the $k$-th part, the feature vectors before and after the $1\times 1$ Conv layer are denoted as $\mathbf{z}_k$ and $\mathbf{f}_k$, respectively. $\mathbf{f}_k$ is utilized as the final part feature and is optimized by the cross-entropy loss:
\begin{equation}
 {\mathcal{L}}_{part}^{id} = \frac{1}{N}\sum_{l=1}^N\sum_{k=1}^K{\mathcal{L}}_{ce}(\mathbf{W}_{k}^{p}{\bf{f}}_{k}^{l}),
\end{equation}
where $\textit{N}$ denotes the batch size. ${\bf{f}}_{k}^{l} \in \mathbb{R}^d$ represents
the $k$-th part-level feature for the $l$-th image in a batch. $\mathbf{W}_{k}^{p}$ stands for the parameters
of the classification layer for the $k$-th part feature. ${\mathcal{L}}_{ce}$ stands for the cross-entropy loss function.

\noindent\textbf{Part Quality Predictor.} \label{Predictor}
A key challenge for occluded ReID is to identify visible body parts.  Recent works~\cite{miao2019PGFA, Gao_2020_CVPR} typically rely on outside tools to infer whether or not one part is visible. However, as argued in Section~\ref{Introduction}, visibility is not equivalent to discriminative power. We accordingly propose an efficient method to predict part feature quality rather than part visibility. As shown in Fig.~\ref{net}, we feed ${\mathbf{z}_k}$ into the part quality predictor module. This module comprises one $1\times 1$ Conv layer, one batch normalization (BN) \cite{57ioffe2015batch} layer, and one sigmoid activation layer. The output of the sigmoid layer is the quality score of the $k$-th part. The parameters of the predictor are not shared between parts.

\noindent\textbf{Joint Optimization.}

As there is no annotation of part quality scores provided, we cannot impose a direct supervision signal on the predicted part quality. Recent works for video-based recognition have revealed that framewise features and quality scores can be jointly optimized with the same identification \cite{zhang2020Multi, zhou2017see} or metric learning loss \cite{liu2017quality}. In these works, e.g., QAN \cite{liu2017quality} and MG-RAFA \cite{zhang2020Multi}, the quality scores are utilized to aggregate multiple frame-level features into a single video-level feature. However, this approach cannot be directly used in image-based ReID, as recognition is based on each single image. Inspired by these works, we propose to jointly optimize part features and part quality scores using triplet loss.
Accordingly, the quality scores in QPM are imposed on part distances instead of frame features. Moreover, both pair-wise part distances and pair part quality scores are included in the triplet loss, instead of using a sample's own quality score only. In this way, our approach optimizes pair-wise part distance and pair part quality scores for occluded ReID in an end-to-end manner.

Specifically, given a probe image $I^p$ and a gallery image $I^g$ to be compared, we first
calculate their part-wise cosine distances $d_k^{pg}$ (1$\leq$ $k$ $\leq$ $K$).
Next, the part-wise distances are summed via weighted average, as follows:
\begin{equation}\label{eq:distance}
D^{pg}_{part} = \frac{\sum_{k=1}^K{{q}_k^p{q}_k^gd_k^{pg}}}{\sum_{k=1}^K{{q}_k^p{q}_k^g}},
\end{equation}
where ${q}_k^p$ and ${q}_k^g$ are the quality score of $k$-th part for probe image $I^p$ and gallery image $I^g$, respectively.  In this way, body parts with low quality scores will contribute less to $D^{pg}_{part}$, which weakens the impact of occlusion.

To sample sufficient
triplets during training, we randomly
sample $A$ images in each of $P$ random identities to
create a mini-batch, the batchsize \emph{N} of which is equal to
$P \times A$. The triplet loss is formulated as $\mathcal{L}^{tp}_{part}=$
\begin{equation} \label{triplet}
\begin{aligned}
\frac{1}{N_{tp}}\sum_{i=1}^P\sum_{a=1}^A
[\alpha&+\max_{p=1...A}D^{a_ip_i}_{part}-
\min_{\substack{n=1...A \\ j=1...P \\ j \neq
i}}D^{a_in_j}_{part}]_+,
\end{aligned}
\end{equation}
where $\alpha$ is the margin of the triplet constraint,
while $N_{tp}$ is the number of triplets in a batch that violate the triplet constraint and $[\cdot]_+$ denotes the hinge loss.
$D^{a_ip_i}_{part}$ and $D^{a_in_j}_{part}$
are calculated by Eq. \ref{eq:distance} and denote the distances of the positive and negative image pairs in a triplet, respectively. 
In order to reduce triplet loss, the part quality predictors have to predict lower scores to occluded parts.

\begin{figure}
    \centerline{\includegraphics[width=0.5\textwidth]{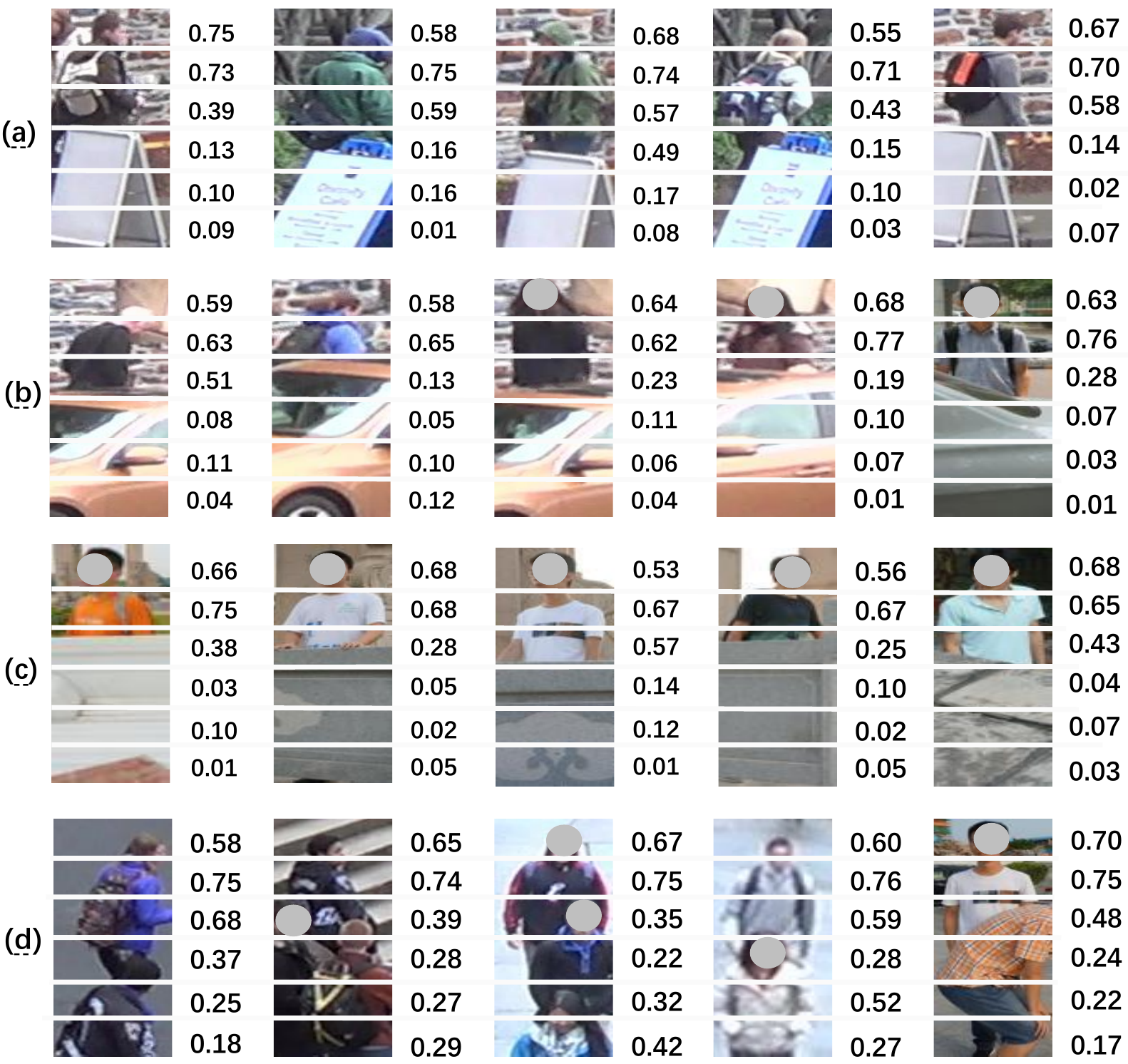}}
     \caption{
    Example images with predicted part quality scores.
    (a) Occlusion caused by billboards. (b) Occlusion caused by cars. (c) Occlusion caused by buildings. (d) Occlusion caused by pedestrians.
     }
    \label{score}
    \end{figure}

\noindent\textbf{\textit{Discussion.}}
With the help of the above constraints, the model can predict the quality scores of the part features without using any external tools. However, this approach still has limitations. Take the predicted quality scores in Fig.~\ref{score} as an example. It can be seen that the model works well for occlusions caused by objects (e.g. cars, trees, and boxes). In these situations, the quality scores of occluded body parts are very low. However, when a pedestrian is occluded by other persons, the model may predict a relatively high score for occluded regions; this is because the model cannot well differentiate body parts of different pedestrians. In the following, we handle the above problem in the global feature learning branch.
Most existing approaches~\cite{miao2019PGFA,Wang_2020_CVPR,zhuo2019novel} simply extract global features from visible regions for each image. However, this strategy suffers from two problems. First, as explained above, visibility or quality prediction of image regions may be interfered by occlusions between pedestrians. Second, as shown in Fig.~\ref{examples1}(b), two images may differ in occlusion locations, meaning that they are not directly comparable even if features are extracted from visible regions for each image. In the following, we propose an Identity-aware Spatial Attention (ISA) approach and an adaptive global feature extraction approach to handle each of these two problems, respectively.

\subsection{Identity-aware Spatial Attention}
\begin{figure}
    \centerline{\includegraphics[width=0.5\textwidth]{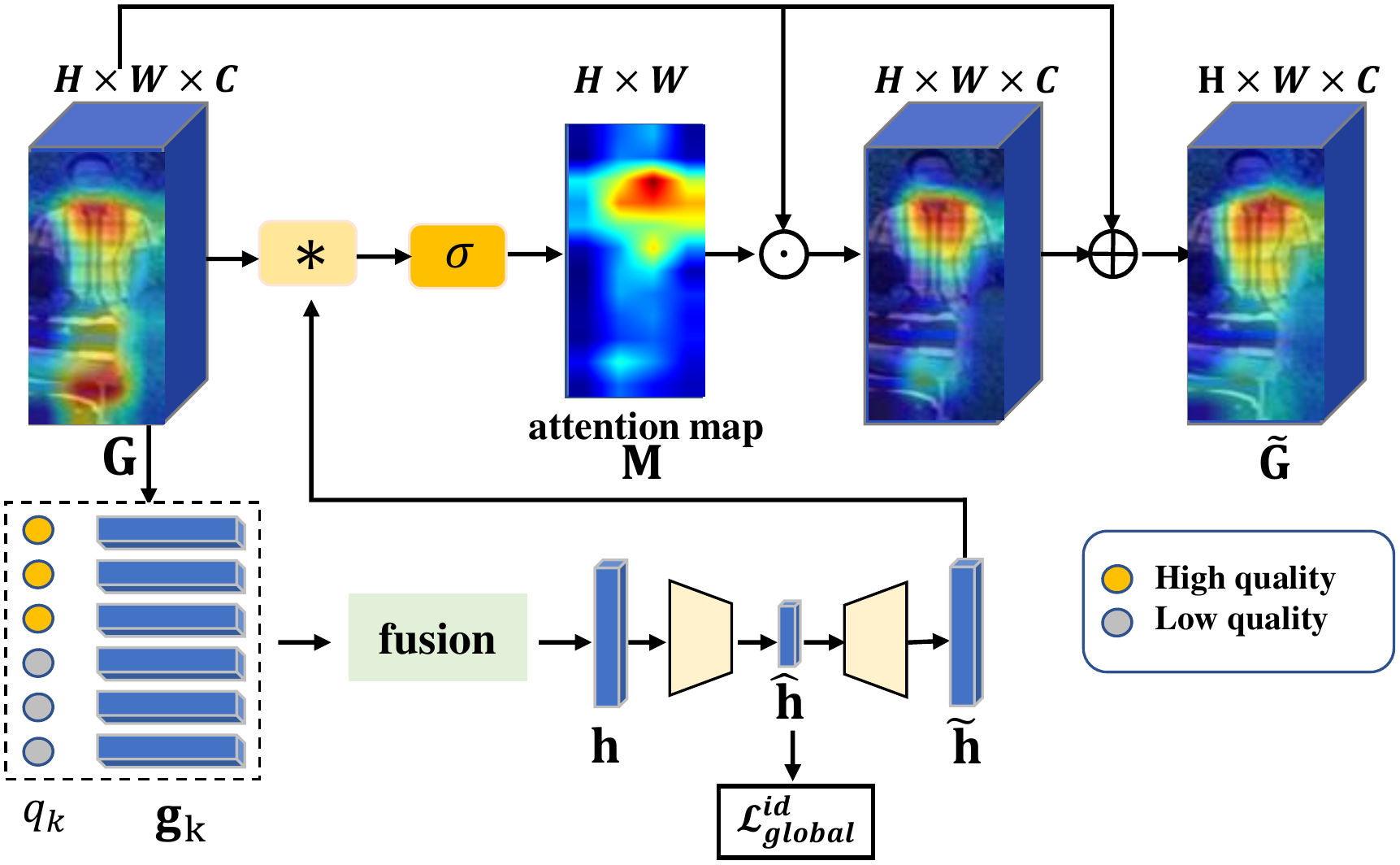}}
     \caption{
     Structure of the Identity-aware Spatial Attention (ISA) module. It first adopts a coarse identity-aware global feature vector $\mathbf{h}$ and a lightweight two-layer network to generate a spatial attention map $\mathbf{M}$. $\mathbf{M}$ is applied on feature maps, i.e. $\mathbf{G}$, to suppress responses of occluded pixels. As $\mathbf{M}$ is identity aware, occlusions between pedestrians can be suppressed.
     (Best viewed in color.)}
    \label{global}
    \end{figure}

Fig.~\ref{global} illustrates the structure of ISA. ISA makes use of a coarse identity-aware feature to generate spatial attention for each image.
This attention suppresses occlusions caused by objects and other pedestrians, meaning that only features of spatial regions relevant to the target pedestrian are highlighted.

In more detail, we feed $\mathbf{F}$ into another $1\times 1$ Conv layer and obtain the feature maps ${\mathbf{G}}$. Global features are then extracted based on $\mathbf{G}$. Similar to the part branch, we uniformly partition $\mathbf{G}$ into $K$ parts and obtain $K$ part-level features that are denoted as $\mathbf{g}_k$ (1$\leq$ $k$ $\leq$ $K$) via RAP operations. We then obtain a coarse identity-aware global feature $\mathbf{h}$ by fusing $\mathbf{g}_k$ via weighted averaging as follows:
\begin{equation}
\label{4}
 \mathbf{h} ={\sum_{k=1}^K{\hat{q}}_k {\mathbf{g}_k}},
\end{equation}
where ${\hat{q}}_k$ is the normalized part quality score produced in the part branch in Section \ref{Predictor}. Formally,
\begin{equation}
   {\hat{q}}_k = \frac{{q}_k}{\sum_{i=1}^K{q}_i}.
\end{equation}
To further reduce the impact of occlusions, we process $\mathbf{h}$ using a simple two-layer network inspired by the squeeze-and-excitation network \cite{hu2018squeeze}. As illustrated in Fig. \ref{global}, the dimension of $\mathbf{h}$ is first reduced via a $1\times 1$ Conv layer which is followed by a ReLU layer, then recovered to the original dimension by means of another $1\times 1$ Conv layer. The reduction ratio of the first Conv layer is set to 4; the output of the two Conv layers are denoted as $\mathbf{\hat{h}}$ and $\mathbf{\widetilde{h}}$, respectively.

Moreover, to ensure that noisy elements in $\mathbf{h}$ are suppressed via the reduction operation, we impose a cross-entropy loss on $\mathbf{\hat{h}}$ as follows:
\begin{equation}
 {\mathcal{L}}_{global}^{id} =  \frac{1}{N}\sum_{l=1}^N{\mathcal{L}}_{ce}(\mathbf{W}^{g}{\mathbf{\hat{h}}}^{l}),
\end{equation}
where ${\mathbf{\hat{h}}}^{l}$ represents the feature produced by the first Conv layer for the $l$-th image in a batch, while $\mathbf{W}^{g}$ denotes the parameters of the classification layers.

By adopting this approach, information in $\mathbf{\widetilde{h}}$ is identity-aware and noise-free. Accordingly, we employ $\mathbf{\widetilde{h}}$ to generate the spatial attention map $\mathbf{M}$ for the feature maps $\mathbf{G}$. Formally:
 \begin{equation}
\mathbf{M} = \sigma(\mathbf{G}\ast \mathbf{\widetilde{h}}),
\end{equation}
where $\sigma$ is a sigmoid function, while $\ast$ represents the inner product between $\mathbf{\widetilde{h}}$ and the feature vector of each pixel in $\mathbf{G}$. Accordingly, $\mathbf{M}$ is a matrix with the same height and width as $\mathbf{G}$.  As identity-relevant pixels obtain a high response value in $\mathbf{M}$, we apply $\mathbf{M}$ to weigh $\mathbf{G}$ and produce new feature maps denoted as $\widetilde{\mathbf{G}}$. The above process can be summarized as follows:
\begin{equation}\label{equ:equ8}
  \widetilde{\mathbf{G}} = \mathbf{M}\odot{\mathbf{G}} + \mathbf{G},
\end{equation}
where $\odot$ signifies the element-wise multiplication between $\mathbf{M}$ and each channel of $\mathbf{G}$.

\noindent\textbf{\textit{Discussion.}}
To the best of our knowledge, ISA is one of the first efficient method to address occlusions between pedestrians in occluded ReID.
A few most recent works \cite{zhao2020not,lu2019see} can potentially solve this problem. These approaches adopt co-attention mechanism and attempt to search for pixel-level correspondence between each pair of images, enabling features to be extracted from semantically corresponding regions.
However, during the training and inference stage, these approaches adopt computationally expensive matrix multiplication to infer semantically corresponding pixels for each pair of query and gallery images. Obviously their computationally cost is significantly higher than that of ISA. Therefore, our method has obvious advantages in efficiency for ReID.

\subsection{Adaptive Global Feature Extraction}
As indicated in Fig.~\ref{vis}, the responses on $\widetilde{\mathbf{G}}$ focus primarily on the body of the pedestrian to identify after the processing of ISA. However,
this does not mean that it is reasonable to extract global-level features directly from  $\widetilde{\mathbf{G}}$;
this is because the two images being compared may differ in terms of their occlusion locations.
To ensure semantic consistency, it is essential to adaptively extract global features from the common non-occluded regions for each image pair.
We designed the Adaptive Global Feature Extraction (AGFE) module with the help of the part quality score to achieve this goal.

For example, given a probe image $I^p$ and a gallery image $I^g$, we first obtain their feature maps ${\widetilde{\mathbf{G}}^p}$ and  ${\widetilde{\mathbf{G}}^g}$ from the output of the ISA module.  In the next step, we equally partition each of them into $K$ parts and apply the RAP operation on each divided feature maps. In this way, we obtain a set of $K$ feature vectors for $I^p$ and $I^g$, denoted as $\widetilde{\mathbf{g}}_k^p$ and $\widetilde{\mathbf{g}}_k^g$ respectively. We then adopt the part quality scores from the part branch to aggregate $\widetilde{\mathbf{g}}_k^p$ and  $\widetilde{\mathbf{g}}_k^g$ and obtain the global-level features $\mathbf{h}^p_g$ and $\mathbf{h}^g_p$ for $I^p$ and $I^g$, respectively. More specifically,
\begin{equation}
\label{p}
 \mathbf{h}^p_g ={\sum_{k=1}^K{\widetilde{q}}_k \widetilde{\mathbf{g}}_k^p},
\end{equation}
\begin{equation}
\label{g}
 \mathbf{h}^g_p ={\sum_{k=1}^K{\widetilde{q}}_k \widetilde{\mathbf{g}}_k^g},
\end{equation}
where ${\widetilde{q}}_k$ denotes the weight for $\widetilde{\mathbf{g}}_k^p$ and  $\widetilde{\mathbf{g}}_k^g$. ${\widetilde{q}}_k$  is computed as follows:
\begin{equation}
   {\widetilde{q}}_k = \frac{{q}^p_k{q}^g_k}{\sum_{i=1}^K{q}^p_i{q}^g_i}.
\end{equation}

The classification loss for the final global representations can thus be formulated as follows:
\begin{equation}
 {\mathcal{L}}_{sg}^{id} =  \frac{1}{NN}\sum_{g=1}^N\sum_{p=1}^N({\mathcal{L}}_{ce}(\mathbf{W}^{s}\mathbf{h}^p_g)+{\mathcal{L}}_{ce}(\mathbf{W}^{s}\mathbf{h}^g_p)).
\end{equation}

We also apply the triplet loss to ensure that the intra-class distances are smaller than the inter-class distances. This triplet loss is similar to Eq.~\ref{triplet}:
\begin{equation}
\begin{aligned}
{\mathcal{L}}^{tp}_{global}=\frac{1}{N_{tp}}\sum_{i=1}^P\sum_{a=1}^A
[\alpha&+\max_{p=1...A}D^{a_ip_i}_{global}-
\min_{\substack{n=1...A \\ j=1...P \\ j \neq
i}}D^{a_in_j}_{global}]_+,
\end{aligned}
\end{equation}
where $D^{a_ip_i}_{global}$ and $D^{a_in_j}_{global}$ represent the cosine distances of global features which are obtained by Eq. \ref{p} and Eq. \ref{g} for the positive and negative image pairs in a triplet, respectively.

Obtaining global features from the common non-occluded regions of each image pair is usually ignored by existing works.
As shown in Table \ref{ablation}, the AGFE module significantly
promotes performance for occluded ReID.
This is because the AGFE module extracts semantically aligned global features.

\subsection{Occluded ReID via QPM}
During training, the overall objective function of QPM can be written as follows:
\begin{equation}
\begin{aligned}
\mathcal{L} = \mathcal{L}_{part}^{id}+\mathcal{L}^{tp}_{part} + \mathcal{L}_{global}^{id} +\mathcal{L}_{sg}^{id} +\mathcal{L}_{global}^{tp}.
\end{aligned}
\end{equation}

During the training process, the overall loss is optimized together. The parameters of the network, including those of the fully connected layers, are optimized together via gradient descent. 

Many existing works design local and global branches \cite{zhang2021information, xie2021pisltrc, miao2019PGFA, Wang_2020_CVPR} for multi-source structural information
integration.
Similar to \cite{miao2019PGFA},
there are two parts in QPM that make up the final distance between one pair of query and gallery images: namely, the distance between part-level features and the distance between global-level features.
In our approach, the distance between the part-level features is computed according to Eq. \ref{eq:distance}. The global-level features with respect to the image pair are obtained according to Eq. \ref{p} and Eq. \ref{g}. Formally,

\begin{equation} \label{distance}
D^{pq} = \gamma D^{pq}_{part} + (1-\gamma)D^{pq}_{global},
\end{equation}
where $\gamma$ is the weight that balances the contributions from $D^{pq}_{part}$ and $D^{pq}_{global}$. $\gamma$ is consistently set to 0.6 in this work.

In the inference stage, we first compute the distance between a query image and each of the gallery images using the part features according to Eq. \ref{eq:distance}. The body parts with low quality scores will contribute less to the distance, which weakens the impact of occlusion.
In this way, we efficiently obtain the top $n$ nearest neighbors for the query image. 
Then, we compute the final distance according to Eq. \ref{distance} between the query image and each of the $n$ nearest neighbors.
Therefore, the AGFE module hardly increases the inference time cost.

\section{Experiments}
       \subsection{Datasets and Settings}
       \noindent{\textbf{Datasets.}} We conduct experiments on four popular databases for occluded ReID, i.e. Partial-iLIDS~\cite{he2018deep}, Partial-REID~\cite{zheng2015partial}, Occluded-Duke~\cite{miao2019PGFA} and  P-DukeMTMC~\cite{zhuo2018occluded}.

   \noindent{\textbf{Partial-iLIDS}}~\cite{he2018deep}
       was constructed based on the iLIDS~\cite{zheng2011person} dataset. It contains 238 images of 119 identities, all of which were captured in an airport.  Some images in the dataset contain people occluded by other individuals or luggages. Each pedestrian has 1 full-body image and 1 occluded image.  All probe images are occluded person images,
while all gallery images are holistic images.

     \noindent{\textbf{Partial-REID}}~\cite{zheng2015partial}
        was collected at a university campus and includes 600 images of 60 pedestrians. Each person has 5 full-body images and 5 occluded images.
These images are collected from different viewpoints, backgrounds, and different types of severe
occlusion. All probe images are occluded person images,
while all gallery images are holistic images.

      \noindent{\textbf{Occluded-Duke}}~\cite{miao2019PGFA}
      was constructed based on the DukeMTMC~\cite{26zheng2017unlabeled} database. It is composed of 15,618 training images of 702 identities, 2,210 occluded query images of 519 identities, and 17,661 gallery images.
      There are rich variations in Occluded-Duke, including different
viewpoints and a large variety of obstacles, including cars,
bicycles, trees, other persons.
      Occluded-Duke is a more difficult and practical dataset since both probe
and gallery images have occlusions.

     \noindent{\textbf{P-DukeMTMC}}~\cite{zhuo2018occluded}
     is another subset of DukeMTMC~\cite{26zheng2017unlabeled}. There are 12,927 training images of 665 identifies, 2,163 query images of 634 identities, and 9,053 gallery images. These images are occluded
by different types of occlusion in public, e.g., people, luggages, cars and guideboards.

    \noindent{\textbf{Implementation Details.}}
    We conduct experiments using the Pytorch framework. We set both $P$ and $A$ to 8; therefore, the batch size is 64. We adopt random erasing to simulate occlusion. All images are resized to $384 \times 128$ pixels and augmented via random horizontal flipping. The number of body parts, i.e. $K$, is set to 6. The margin $\alpha$ for the triplet loss is set to 0.3. The number of the nearest neighbors, i.e., $n$, is set as 30.
    The SGD optimizer is utilized for model optimization.
     {Following \cite{sun2018beyond,ding2020multi,03_wang2018learning}, we do not use weight regularization in the SGD optimizer.} Fine-tuned from the IDE model \cite{19zheng2017person}, the QPM is trained in an end-to-end fashion for 70 epochs. The initialized learning rate is set to 0.01 and is reduced  by multiplying 0.1 for every 20 epochs.

   \noindent{\textbf{Evaluation Protocols.}}
We report the Cumulated Matching Characteristics (CMC) and mean Average Precision (mAP) value for the proposed approach. The evaluation package is provided by \cite{zhou2019torchreid}, and all the experimental results are obtained in the single query setting.

Moreover, we provide stability analysis on the performance of QPM in the supplementary material.

    \begin{table}
  \renewcommand\arraystretch{1}
     \caption{Performance comparisons on Occluded-Duke under supervised setting. $\ast$ indicates that a different backbone is used. $+$ represents the extended version of the conference paper.}
     \label{Occluded-Duke}
     \centering
     {
     \begin{tabular}{c|cccc}
    \toprule[1pt]
      Method  &Rank-1 & Rank-5 & Rank-10 & mAP    \\

      \hline
      \hline
      Part-Aligned~\cite{43zhao2017deeply} &28.8 &44.6 &51.0 &20.2 \\
      Part-Bilinear~\cite{suh2018part} &36.9 &- &- &- \\
      Random Erasing~\cite{38zhong2017random} &40.5 &59.6 &66.8 &30.0 \\
      DSR~\cite{he2018deep} &40.8 &58.2 &65.2 &30.4 \\
      SFR~\cite{he2018recognizing} &42.3 &60.3 &67.3 &32.0 \\
      PCB~\cite{sun2018beyond} &42.6 &57.1 &62.9 &33.7 \\
      Adver Occluded~\cite{huang2018adversarially} &44.5 &- &- &32.2 \\
PVPM~\cite{Gao_2020_CVPR} &47.0 &- &- &37.7  \\
      PGFA~\cite{miao2019PGFA} &51.4 &68.6 &74.9 &37.3 \\
      HOReID\cite{Wang_2020_CVPR} &55.1 &- &- &43.8 \\
      $\rm{PGFA}^{+}$ \cite{miao2021identifying} &56.3 &72.4 &78.0 &43.5 \\
      $\rm{ISP}^{*}$ \cite{zhu2020identity} &62.8 &78.1 &82.9 &52.3 \\

      \hline
    \bfseries QPM &\bfseries64.4 &\bfseries79.3 &\bfseries84.2 &\bfseries49.7 \\
    \bfseries $\rm {QPM}^*$&\bfseries66.7 &\bfseries80.2 &\bfseries84.4 &\bfseries53.3 \\

      \toprule[1pt]
      \end{tabular}
     \vspace{-3mm}
     }
     \end{table}

        \begin{table}
             \caption{Performance comparisons on P-DukeMTMC under supervised setting.}
             \centering
             \label{Psupervised}
             \scalebox{.93}
             {\begin{tabular}{c|cccc}
            \toprule[1pt]
             Method & Rank-1 & Rank-5 &Rank-10 &mAP \\
              \hline
                          \hline

             Teacher-S~\cite{zhuo2019novel} &51.4 &50.9 &- &-\\
            PCB~\cite{sun2018beyond} &79.4 &87.1 &90.0 &63.9      \\
             IDE~\cite{19zheng2017person} &82.9 &89.4 &91.5 &65.9\\
             PVPM~\cite{Gao_2020_CVPR} &85.1 &91.3 &93.3 &69.9   \\
              { $\rm{PGFA}^{+}$ \cite{miao2021identifying}}  &  {85.7}	&  {92.0}	&  {94.2}	&  {72.4} \\
                 { $\rm{ISP}^{*}$ \cite{zhu2020identity}} &  {89.0} &  {94.1} &  {95.3} &  {74.7} \\
            \hline
         \bfseries QPM &\bfseries89.4 &\bfseries93.9  &\bfseries95.6 &\bfseries74.4\\
         \bfseries  {$\rm {QPM}^*$} &\bfseries {90.7} &\bfseries   {94.4}  &\bfseries   {95.9} &\bfseries   {75.3}\\

            \toprule[1pt]
             \end{tabular}
             }
          \end{table}

      \subsection{Performance under Supervised Setting}

    For the two large-scale datasets, i.e. Occluded-Duke \cite{miao2019PGFA}  and P-DukeMTMC \cite{zhuo2018occluded}, we train QPM using their own training sets, respectively.

    \noindent{\textbf{Results on Occluded-Duke.}}
The performance of QPM and state-of-the-art methods on Occluded-Duke are tabulated in Table \ref{Occluded-Duke}. Some recent methods have achieved competitive performance with the help of pose landmarks: for example, HOReID \cite{Wang_2020_CVPR} learns high-order relation and topology information for local features of visible landmarks, facilitating better match between probe and gallery images, and achieves 55.1\% Rank-1 accuracy and 43.8\% mAP. In comparison, QPM significantly outperforms HOReID by 9.3\% and 5.9\% in terms of Rank-1 accuracy and mAP, respectively. Moreover, QPM does not depend on pose estimation tools in either training or testing. This remarkable performance improvement clearly demonstrates the effectiveness of QPM.

One recent method $\rm{ISP}^{*}$ \cite{zhu2020identity} achieves strong performance with a deeper backbone model, i.e. HRNet-W32 \cite{sun2019deep}.
With the same backbone model, $\rm QPM^{*}$ achieves higher Rank-1 and mAP performance than $\rm{ISP}^{*}$.

\noindent{\textbf{Results on P-DukeMTMC.}}
Comparison results on P-DukeMTMC are summarized in Table \ref{Psupervised}. As the table shows, QPM achieves 89.4\% Rank-1 accuracy and 74.4\% mAP, surpassing the best previous method PVPM \cite{Gao_2020_CVPR} by 4.3\% and 4.5\% in terms of Rank-1 accuracy and mAP, respectively. The above comparison results are consistent with those obtained on the Occluded-Duke database.
These experimental results further demonstrate that our method can effectively solve the occlusion problem for ReID.

    \subsection{Performance under Transfer Setting}\label{sec:transfer}
       Following~\cite{Gao_2020_CVPR,Wang_2020_CVPR,sun2019perceive,he2018deep}, ReID model in this setting are trained using the Market-1501 database~\cite{zheng2015scalable}. Then, the model is directly evaluated on the Partial-REID, Partial-iLIDS, and P-DukeMTMC databases.

    \noindent{\textbf{Results on Partial-REID and Partial-iLIDS.}}
    Each of the two databases contain two types of testing data: namely, partial images from which occluded regions have been manually removed, and the original occluded images. Similarly, depending on the testing data used, existing methods can be roughly divided into two categories: those using partial images and those using the original occluded images. The performance of QPM and state-of-the-art methods are tabulated in Table \ref{Partial}.

As is evident from the table, for Partial-iLIDS, it can be seen that QPM outperforms all other methods that also evaluate on the original occluded images. For example, QPM beats PGFA~\cite{miao2019PGFA} by 8.2\% and 4.8\% in terms of Rank-1 and Rank-3 accuracy respectively. It also outperforms one of the most recent methods using partial data, i.e. HOReID~\cite{Wang_2020_CVPR}, by 4.7\% in terms of Rank-1 accuracy. For Partial-REID, QPM outperforms all other methods using the original occluded images: for example, QPM beats PVPM~\cite{Gao_2020_CVPR} by 3.4\% in terms of Rank-1 accuracy. While its performance is slightly lower than HOReID~\cite{Wang_2020_CVPR} (which evaluates on partial data), it should be noted that unlike HOReID, our method does not require either the manual removal of occluded regions during testing or any additional tools. 
  {$\rm{ISP}^{*}$ \cite{zhu2020identity} achieves 65.7\% Rank-1 and 75.3\% Rank-3 accuracy on Partial-REID, which is lower than those by QPM. This is because the image quality of Partial-REID is poor and the image resolution is relatively low. In this case, ISP is prone to errors in pixel classification. In comparison, our method is more robust.}

\begin{table}
\centering
\caption{Performance comparisons on Partial-iLIDS and Partial-REID under transfer setting. $+$ represents the extended version of the conference paper.}
\begin{tabular}{c|c|cc|cc}
\toprule[1pt]
  \multicolumn{2}{c|}{\multirow{2}{*}{Methods}} & \multicolumn{2}{c|}{Partial-iLIDS} & \multicolumn{2}{c}{Partial-REID} \\ \cline{3-6}
  \multicolumn{2}{c|}{} &R-1  &R-3   & R-1 &R-3 \\
\hline
\hline
  \multirow{8}*{\rotatebox{90}{partial}}

  & MTRC \cite{liao2012partial} & 17.7 & 26.1 & 23.7 & 27.3  \\
      & AMC+SWM\cite{zheng2015partial} & 21.0 & 32.8 & 37.3 & 46.0  \\
      &DSR \cite{he2018deep} & 58.8 & 67.2 & 50.7 & 70.0 \\
      &DCR \cite{gao2020dcr}  &60.5 &69.7 &52.0 &67.5  \\
      &STNReID \cite{luo2020stnreid}  &54.6 &71.3 &66.7 &80.3  \\
    &  SFR \cite{he2018recognizing} & 63.9 & 74.8 & 56.9 & 78.5\\
     & VPM \cite{sun2019perceive} & 65.5 & 74.8 & 67.7 & 81.9\\
     & HOReID \cite{Wang_2020_CVPR}       & 72.6 & 86.4 & 85.3 & 91.0\\
\hline
\multirow{6}*{\rotatebox{90}{original}}

 & MaskReID \cite{qi2018maskreid} & 33.0 & - & 28.7 & -\\
    &  PCB \cite{sun2018beyond} & 46.8 &- & 56.3 & - \\
    &  { $\rm{ISP}^{*}$ \cite{zhu2020identity}} &  {66.4} &  {80.7} &  {65.7} &  {75.3} \\
    &  PGFA \cite{miao2019PGFA} & 69.1 & 80.9 &68.0 &80.0\\
    & PVPM~\cite{Gao_2020_CVPR} &- &- &78.3 &-\\
    & $\rm{PGFA}^{+}$ \cite{miao2021identifying}  &70.6 &81.3 &72.5 &83.0 \\
\cline{2-6}
  &\bfseries QPM &\bfseries77.3 &\bfseries85.7  &\bfseries81.7 &\bfseries88.0 \\
    \toprule[1pt]
\end{tabular}
\label{Partial}
\end{table}


    \noindent{\textbf{Results on P-DukeMTMC.}}
Finally, we evaluate the performance
of QPM on P-DukeMTMC under transfer setting. The results presented in Table \ref{P-Duke} show that QPM achieves state-of-the-art performance under all metrics. For example, QPM outperforms one of the most recent methods, i.e. PVPM~\cite{Gao_2020_CVPR}, by 5.8\% and 1.9\% in terms of Rank-1 accuracy and mAP, respectively.
Experiments on this database further justifies the effectiveness of QPM.

     \begin{table}
     \caption{Performance comparisons on P-DukeMTMC under transfer setting.}
     \label{P-Duke}
     \centering
     {
     \begin{tabular}{c|cccc}
    \toprule[1pt]
      Methods & Rank-1 & Rank-5 & Rank-10 & mAP \\
      \hline
      \hline
      HACNN~\cite{06_li2018harmonious} &30.4 &42.1 &49.0 &17.0\\
      MLFN~\cite{45changCVPR2018multi} &31.3 &43.6 &49.6 &18.1\\
      OsNet~\cite{zhou2019omni} &33.7 &46.5 &54.0 &20.1\\
      IDE~\cite{19zheng2017person} &36.0 &49.3 &55.2 &19.7 \\
      Part Bilinear~\cite{suh2018part} &39.2 &50.6 &56.4 &25.4\\
      PCB~\cite{sun2018beyond} &43.6 &57.1 &63.3 &24.7\\
      PGFA~\cite{miao2019PGFA} &44.2 &56.7 &63.0 &23.1\\
         { $\rm{ISP}^{*}$ \cite{zhu2020identity}} &  {46.3} &  {56.9} &  {60.8} &  {26.4} \\
         { $\rm{PGFA}^{+}$ \cite{miao2021identifying}}  &  {48.2}	&  {59.6}	&  {65.8}	&  {26.8} \\
      PVPM~\cite{Gao_2020_CVPR} &51.5 &64.4 &69.6 &29.2\\
      \hline
      \bfseries QPM &\bfseries57.3 &\bfseries69.9 &\bfseries75.5 &\bfseries31.1 \\
    \toprule[1pt]
      \end{tabular}
     }
     \end{table}

\subsection{Ablation Study}
Ablation study are conducted on two large-scale datasets, i.e. Occluded-Duke~\cite{miao2019PGFA} and  P-DukeMTMC~\cite{zhuo2018occluded}.
The experimental results on the reported two benchmarks are shown in Table \ref{ablation}. It should be noted that Table \ref{ablation} lists the results on Occluded-Duke under supervised setting and the results on P-DukeMTMC under transfer setting. These results show the robustness and effectiveness of our method under different experimental settings.

\begin{equation}
\begin{aligned}
\mathcal{L} = \mathcal{L}_{part}^{id}+\mathcal{L}^{tp}_{part} + \mathcal{L}_{global}^{id} +\mathcal{L}_{sg}^{id} +\mathcal{L}_{global}^{tp}.
\end{aligned}
\end{equation}

\noindent\textbf{Effectiveness of the quality scores.}
In Table \ref{ablation}, `Baseline' refers to the PCB model \cite{sun2018beyond}. `Baseline(+triplet)' equips PCB with the triplet loss in Eq. \ref{triplet}, while all part quality scores are set to 1. `Part branch' means that we adopt the part feature learning branch only in QPM for ReID. 
As shown in Table \ref{ablation}, `Baseline(+triplet)' slightly improves the performance relative to the baseline. In comparison, the `Part branch' brings in significant performance promotion, suggesting that part quality scores considerably  benefit the occluded ReID task.

\noindent\textbf{Effectiveness of the ISA module.}
In Table \ref{ablation}, `GAP global' means that we perform GAP on the feature maps $\mathbf{G}$ to obtain the global feature for each image. 
`AGFE global' means that we utilize adaptive global feature extraction module.
When the ISA module is equipped, performance of both types of global features is promoted. In particular, ISA promotes the Rank-1 accuracy of `AGFE global' by 6.0\% and 4.6\%, as well as mAP by 3.0\% and 1.8\%, on the two databases, respectively.

\noindent\textbf{Effectiveness of the AGFE module.}
Compared with `GAP global' and `GAP global(+ISA)', the AGFE module consistently brings about significant performance gains. For example, `AGFE global(+ISA)' outperforms `GAP global(+ISA)' in the Rank-1 accuracy by as much as 19.9\% on Occluded-Duke. These experimental results  indicate that it is vital to adaptively extract global features from the common non-occluded regions for each image pair.

Moreover, we compare the performance of `AGFE global' with `SI global' in Table \ref{Global}. `SI global' means that we obtain global features for each single image (SI) using Eq. \ref{4}, without considering the difference in occlusion locations for each image pair. To facilitate a fair comparison, `SI global' adopts the same loss functions as `AGFE global', i.e., the cross-entropy loss and triplet loss. It is shown that the performance of `SI global' drops dramatically compared with `AGFE global', which suggests that Eq. \ref{4} alone cannot promote ReID performance. This experimental result indicates that it is vital to extract semantically consistent global features for each image pair.

\noindent\textbf{Effectiveness of the combination.}
With both ISA and AGFE modules, the quality of global features is promoted significantly. For example, `AGFE global(+ISA)' outperforms `GAP global' in the Rank-1 accuracy by as much as 21.8\% and 20.0\% on Occluded-Duke and P-DukeMTMC databases, respectively.  

 Finally, the combination of the part branch and the global branch, which is denoted as QPM in Table \ref{ablation}, achieves better performance than using either one branch alone. The above comparisons justify the effectiveness of each key component in QPM.

\begin{table}

     \caption{Ablation study on each key component of QPM.}
    \label{ablation}
     \centering
     {
     \begin{tabular}{c|cc|cc}
    \toprule[1pt]
      \multirow{2}{*}{Method}  &\multicolumn{2}{c|}{Occluded-Duke}&\multicolumn{2}{c}{P-DukeMTMC}\\
          & Rank-1  & mAP  & Rank-1  & mAP     \\
      \hline
      \hline
      Baseline &42.1  &34.8 &47.2  &22.9 \\
      Baseline(+triplet) &43.7  &37.0 &48.5  &26.4 \\
      Part branch   &58.3  &46.7 &55.1  &28.8 \\
      \hline
      GAP global & 40.8  & 31.7  & 26.0  & 14.7 \\
      GAP global(+ISA) & 42.7  & 33.0 & 27.0  & 15.5 \\
      AGFE global &56.6  &42.2 &41.4  &23.3 \\
     AGFE global(+ISA)  &62.6  &45.2 &46.0  &25.1 \\
         \hline
      \bfseries QPM &\bfseries64.4  &\bfseries49.7 &\bfseries57.3  &\bfseries31.1 \\
    \toprule[1pt]
      \end{tabular}
     }
     \end{table}

\begin{table}

     \caption{Performance comparison between different types of global features.}
    \label{Global}
     \centering
     {
     \begin{tabular}{c|cc|cc}
    \toprule[1pt]
      \multirow{2}{*}{Method}  &\multicolumn{2}{c|}{Occluded-Duke}&\multicolumn{2}{c}{P-DukeMTMC}\\
          & Rank-1  & mAP  & Rank-1  & mAP     \\
      \hline
      \hline
      GAP global & 40.8  & 31.7  & 26.0  & 14.7 \\
      SI global &38.5  &26.6 &26.5 &13.1 \\
      AGFE global &56.6  &42.2 &41.4  &23.3 \\
    \toprule[1pt]
      \end{tabular}
     }
     \end{table}

\begin{table}
\centering
\caption{Performance comparisons of different attention models.}
     \label{sa}

\begin{tabular}{c|cc|cc}
\toprule[1pt]
  \multirow{2}{*}{Methods} & \multicolumn{2}{c|}{Occluded-Duke} & \multicolumn{2}{c}{P-DukeMTMC} \\
  &Rank-1  & mAP  & Rank-1 & mAP \\
      \hline
      \hline
        AGFE global  &56.6  &42.2  & 41.4 & 23.3  \\
         \hline
         +FC-SA \cite{liu2019spatial} & 56.9 & 42.0 & 42.8 & 23.8  \\
        +RGA-SA \cite{zhang2020relation} & 57.6 & 41.2 & 40.9 & 23.1  \\
        +RA-SA \cite{wang2017residual} & 58.1 & 42.7 & 41.3 & 23.3  \\
      +CBAM-SA \cite{woo2018cbam} & 59.3 & 43.9 & 43.1 & 24.5 \\
      \hline
      \bfseries+ISA &\bfseries62.6 &\bfseries45.2  &\bfseries46.0 &\bfseries25.1 \\
    \toprule[1pt]
      \end{tabular}

     \end{table}
\noindent\textbf{ISA vs. Other Approaches.}
To facilitate fair comparison, all experiments are based on the `AGFE global' model. We equip the `AGFE global' model with different spatial attention modules respectively and summarize their performance in Table \ref{sa}. It is shown that ISA significantly outperforms all other methods by at least 3.3\% and 2.9\% in terms of Rank-1 accuracy on Occluded-Duke and P-DukeMTMC, respectively. This is because ISA is identity-aware; therefore, it effectively handles the occlusions between pedestrians.

\begin{figure}
    \centerline{\includegraphics[width=0.5\textwidth]{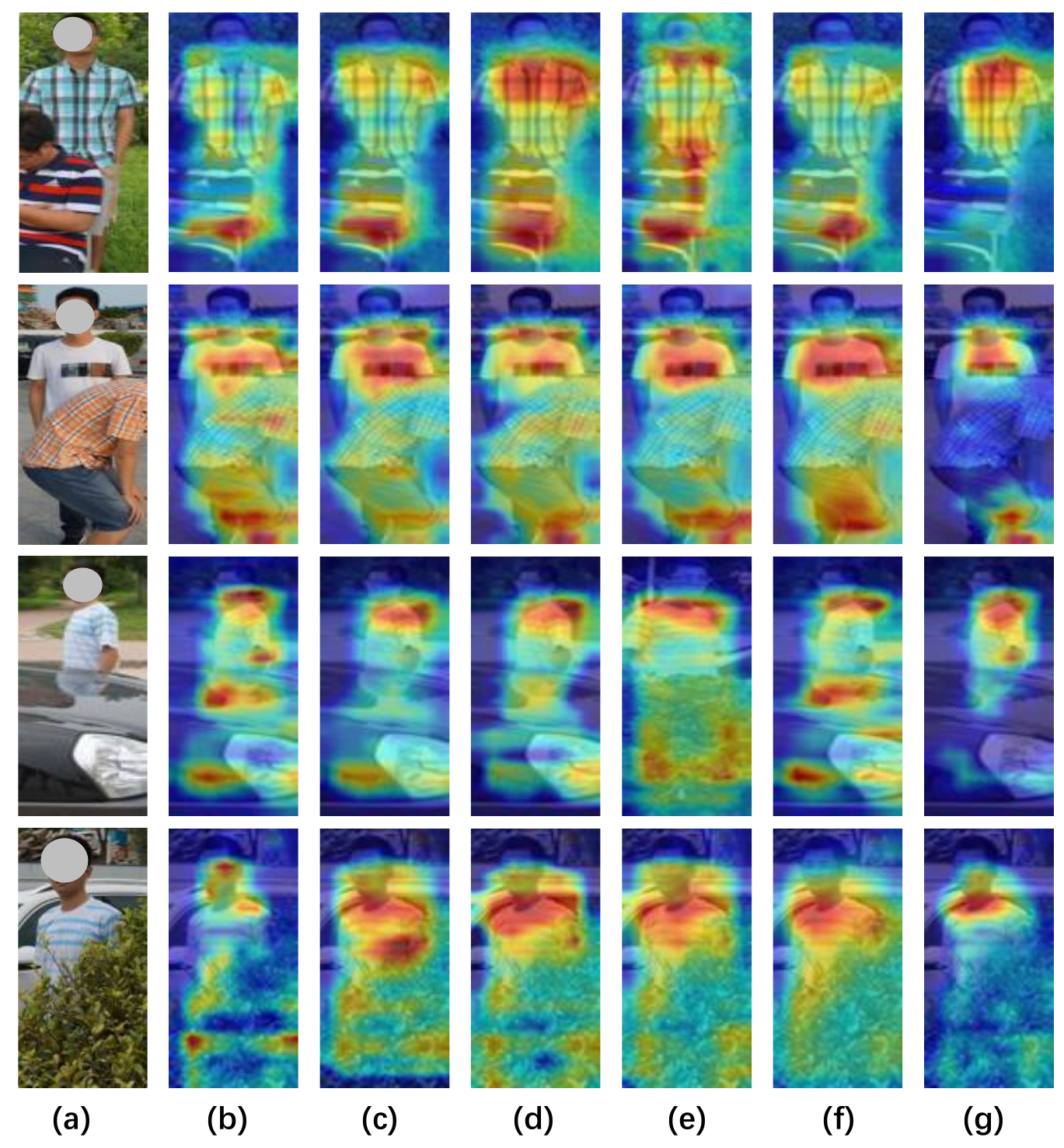}}
     \caption{
     Illustration of heat maps for feature maps after processing by different attention models. (a) The original image. (b) w/o attention. (c) FC-SA. (d) RGA-SA. (e) RA-SA. (f) CBAM-SA. (g) ISA (ours). (Best viewed in color.)
     }
    \label{vis}
    \end{figure}

Moreover, we illustrate the heat maps for feature maps after the processing of different attention modules in Fig.~\ref{vis}. We have the following observations. First, heat maps for the baseline model have high responses on both occluded and non-occluded regions; second, existing popular spatial attention models~\cite{liu2019spatial,zhang2020relation,wang2017residual,woo2018cbam} handle occlusions between pedestrians poorly; third, with the identity-aware guidance, our ISA module can well differentiate discriminative body parts from the occluded ones by both objects and other pedestrians. The above comparisons further demonstrate the effectiveness of ISA.

   \begin{table}
  \renewcommand\arraystretch{1}
     \caption{Comparisons of model complexity on the Occluded-Duke database.
$+$ represents the extended version of the conference paper.}
     \label{efficiency}
     \centering
     {
     \begin{tabular}{c|cccc}
    \toprule[1pt]
      Method  &Train & Inference & Params & Rank-1    \\
      \hline
      \hline
      PGFA~\cite{miao2019PGFA} &0.82s/iter &0.78s/img &~115.5M &51.4 \\
      HOReID \cite{Wang_2020_CVPR} &1.34s/iter &1.94s/img &~117.5M &55.1 \\
      $\rm{PGFA}^{+}$ \cite{miao2021identifying} &0.88s/iter &0.79s/img &~115.5M &56.3 \\
      \hline
    \bfseries QPM &\bfseries0.68s/iter &\bfseries0.47s/img &\bfseries~29.0M &\bfseries64.4 \\
      \toprule[1pt]
      \end{tabular}
     }
     \end{table}

\noindent\textbf{Comparisons of Model Complexity.}
In this experiment, we demonstrate that QPM not only
achieves superior performance in terms of occluded ReID accuracy,
but also offers advantages in terms of both its time and
space complexities.

Three recent powerful occluded ReID approaches
are compared: PGFA~\cite{miao2019PGFA}, HOReID \cite{Wang_2020_CVPR}, and $\rm{PGFA}^{+}$ \cite{miao2021identifying}. 
To facilitate fair comparison, all experimental settings are consistent with the paper description.
Following \cite{miao2019PGFA,Wang_2020_CVPR,miao2021identifying}, the input images are resized to 256 × 128 pixels for HOReID  and 384 $\times$ 128 pixels for PGFA, $\rm{PGFA}^{+}$ and QPM. 
The batch size is set to 64 uniformly for all methods.
Comparisons are conducted on a Titan V GPU, and results are summarized
in Table \ref{efficiency}. 
The inference time cost in Table \ref{efficiency} includes the feature extraction of the query image and the matching time of all gallery images.

As is evident from the Table \ref{efficiency}, the model size of QPM is much smaller since it does not need an additional detection model. In addition, QPM has a faster training and testing speed.
Specifically, our test time is only 60\% and 24\% of PGFA~\cite{miao2019PGFA} and HOReID \cite{Wang_2020_CVPR}, respectively. 
This is because QPM does not require time-consuming human key point extraction.
Although QPM uses less additional information during training and testing than PGFA~\cite{miao2019PGFA} and HOReID \cite{Wang_2020_CVPR}, it still has significant performance advantages.
Accordingly, the above comparisons demonstrate that
the proposed QPM model is both compact and efficient.

\subsection{Parameter Analysis}
\noindent\textbf{The Impact of Part Number $K$.}
In this experiment, we analyze the impact of the part number $K$. Experimental results are summarized in Table~\ref{k}. It is shown that the optimal value of $K$ is 6, which is consistent with the conclusion in~\cite{sun2018beyond}. Therefore, we consistently set $K$ to 6 in this work.

\begin{table}
\centering
     \caption{Performance of QPM with different values of $K$.}
     \label{k}
     \centering
     {
     \begin{tabular}{c|cc|cc}
    \toprule[1pt]
      \multirow{2}{*}{Part number $K$}  &\multicolumn{2}{c|}{Occluded-Duke}&\multicolumn{2}{c}{P-DukeMTMC}\\\cline{2-5}
          & Rank-1  & mAP  & Rank-1  & mAP     \\
      \hline
      \hline
         2 &61.7  &49.1 &44.2  &25.6 \\
      3 &61.9 &49.4 &49.1 &26.8\\
      4 &62.7  &49.4 &50.5  &28.4 \\
      \bfseries 6  & \bfseries64.4  & \bfseries49.7 & \bfseries57.3  & \bfseries31.1 \\
      8  &62.7  &49.5 &53.7  &28.8 \\
    \toprule[1pt]
      \end{tabular}
     }
     \end{table}

\begin{table}

  \renewcommand\arraystretch{1}
     \caption{Performance of QPM with different values of $d$.}
     \label{d}
     \centering
     {
     \begin{tabular}{c|cc|cc}
    \toprule[1pt]
      \multirow{2}{*}{Feature dimension $d$}  &\multicolumn{2}{c|}{Occluded-Duke}&\multicolumn{2}{c}{P-DukeMTMC}\\\cline{2-5}
          & rank-1  & mAP  & rank-1  & mAP     \\
      \hline
      \hline

     128 &61.0  &48.7 &52.0  &28.8 \\
      256 &62.9 &49.5 &55.2 &29.3\\
      512 &63.5  &49.7 &56.3  &30.8 \\
      \bfseries1024  & \bfseries64.4  & \bfseries49.7 & \bfseries57.3  & \bfseries31.1 \\

    \toprule[1pt]
      \end{tabular}
     }
     \end{table}
     
    \begin{table}

  \renewcommand\arraystretch{1}
     \caption{  {Performance of QPM with different values of $\gamma$.}}
     \label{gamma}
     \centering
     {
     \begin{tabular}{c|cc|cc}
    \toprule[1pt]
   \multirow{2}{*}{Weight $\gamma$}  &\multicolumn{2}{c|}{Occluded-Duke}&\multicolumn{2}{c}{P-DukeMTMC}\\\cline{2-5}
          & rank-1  & mAP  & rank-1  & mAP     \\
      \hline
      \hline

     0.5 &64.0  &49.3 &57.0  &30.7 \\
        \bfseries0.6  & \bfseries64.4  & \bfseries49.7 & \bfseries57.3  & \bfseries31.1 \\
      0.7 &63.6  &49.3 &57.1  &30.6 \\
            0.8 &62.9  &49.3 &56.6  &30.1 \\
            0.9 &61.4  &48.7 &56.2  &29.6 \\
            1.0  &58.3  &46.7 &55.1  &28.8 \\

    \toprule[1pt]
      \end{tabular}
     }
     \end{table}

\noindent\textbf{The Impact of the Feature Dimension $d$.}
Table \ref{d} shows the ReID performance with different feature dimension $d$. QPM consistently achieves state-of-the-art performance when $d$ is set to 256, 512, and 1024, with the best result achieved when $d$ is set to 1024.

  {\noindent\textbf{The Impact of the weight $\gamma$.}}
  {Table \ref{gamma} shows the ReID performance with different values of $\gamma$. It is shown that the optimal value of $\gamma$ is 0.6. Therefore, we consistently set $\gamma$ to 0.6 in this work.}

\section{Conclusion}

In this paper, we propose a novel framework named QPM to handle the occluded person ReID problem. Unlike most existing methods, which depend on visibility cues from outside tools, QPM jointly learns part features and predicts part quality in an end-to-end framework without using any annotations or outside tools. Moreover, based on the predicted part quality scores, we propose a novel identity-aware spatial attention (ISA) model to handle occlusion between pedestrians. We further design a novel approach that adaptively generates global features from common non-occluded regions for each image pair. Finally, extensive experiments on four popular datasets demonstrate the effectiveness of QPM.

\bibliographystyle{IEEEtran}
\bibliography{egbib}

\end{document}